\documentclass[letterpaper]{article} 
\usepackage{aaai25}  
\usepackage{times}  
\usepackage{helvet}  
\usepackage{courier}  
\usepackage[hyphens]{url}  
\usepackage{graphicx} 
\usepackage{natbib}  
\usepackage{caption} 
\usepackage{algorithm}
\usepackage{algorithmic}

\usepackage{subfig} 
\usepackage{subfloat}

\usepackage{makecell}
\usepackage{color}
\usepackage{textcomp}
\usepackage{url}
\usepackage{verbatim}
\usepackage{graphicx}
\usepackage{amsmath,amsfonts}
\usepackage{array}
\usepackage{multirow}
\usepackage{booktabs}
\usepackage{float}

\usepackage{newfloat}
\usepackage{listings}
\DeclareCaptionStyle{ruled}{labelfont=normalfont,labelsep=colon,strut=off} 
\floatstyle{ruled}
\newfloat{listing}{tb}{lst}{}
\floatname{listing}{Listing}
\title{HEROS-GAN: Honed-Energy Regularized and Optimal Supervised GAN \\for Enhancing Accuracy and Range of Low-Cost Accelerometers
}

\author{
	Yifeng Wang, Yi Zhao\thanks{Corresponding author.}
}
\affiliations{
	School of Science, Harbin Institute of Technology, Shenzhen\\
	wangyifeng@stu.hit.edu.cn, zhao.yi@hit.edu.cn
}

\usepackage{bibentry}

\begin{document}

\maketitle

\begin{abstract}
Low-cost accelerometers play a crucial role in modern society due to their advantages of small size, ease of integration, wearability, and mass production, making them widely applicable in automotive systems, aerospace, and wearable technology.
However, this widely used sensor suffers from severe accuracy and range limitations. To this end, we propose a honed-energy regularized and optimal supervised GAN (HEROS-GAN), which transforms low-cost sensor signals into high-cost equivalents, thereby overcoming the precision and range limitations of low-cost accelerometers. 
Due to the lack of frame-level paired low-cost and high-cost signals for training, we propose an Optimal Transport Supervision (OTS), which leverages optimal transport theory to explore potential consistency between unpaired data, thereby maximizing supervisory information. Moreover, we propose a Modulated Laplace Energy (MLE), which injects appropriate energy into the generator to encourage it to break range limitations, enhance local changes, and enrich signal details. 
Given the absence of a dedicated dataset, we specifically establish a Low-cost Accelerometer Signal Enhancement Dataset (LASED) containing tens of thousands of samples, which is the first dataset serving to improve the accuracy and range of accelerometers and is released in Github.
Experimental results demonstrate that a GAN combined with either OTS or MLE alone can surpass the previous signal enhancement SOTA methods by an order of magnitude. Integrating both OTS and MLE, the HEROS-GAN achieves remarkable results, which doubles the accelerometer range while reducing signal noise by two orders of magnitude, establishing a benchmark in the accelerometer signal processing.

\end{abstract}

%

\section{Introduction}

Low-cost accelerometers are indispensable in modern technology due to their small size, ease of integration, and widespread availability \cite{9500152,ehatisham2021expert}. In industry, they are used for posture control, navigation, and path planning of machinery. By measuring motion, accelerometers enable the monitoring and control of equipment in complex environments, aiding in structural health monitoring and vehicle dynamic performance testing \cite{caesar2020nuscenes}. 
On production lines, accelerometers detect abnormal vibrations in real time to predict equipment failures. In smart manufacturing, they assist in positioning and motion control, enhancing product quality and stability \cite{wang2024wavelet}. In robotics, accelerometers monitor robotic arm movements for precise operations \cite{8794399,liu2020wearable}. In IoT applications, they ensure optimal operation by monitoring equipment status \cite{yang2023lidar}.
In communication, accelerometers are used in mobile devices for posture detection and user interaction \cite{liu2020tlio,li2023attitude}. They enable automatic screen rotation, step counting, game control, and AR applications, enhancing user convenience and interactivity. In smartphones, accelerometers support gesture recognition, fall detection, and gait analysis, offering intelligent and personalized services \cite{10318077}. In medical and health monitoring, accelerometers have promising applications. Embedded in wearable devices, they allow real-time monitoring of patients' activity and physiological parameters, aiding diagnosis and health management. For example, accelerometers detect falls in the elderly by monitoring sudden changes in body acceleration and providing emergency alerts. They also monitor movements in rehabilitation training, assess recovery progress, and guide exercises.

However, the widely used low-cost accelerometers face significant accuracy and range issues \cite{malayappan2022sensing}. Limited accuracy with severe noise hinders low-cost  accelerometers in high-precision motion capture and subtle vibration monitoring \cite{10614868}. In industrial automation, precise machine motion control relies on high-quality acceleration signals, but severe noise can significantly affect the stable operation and control of machines. In medical monitoring, noise interferes with accurately measuring movements and physiological parameters, hindering remote diagnosis and health management \cite{gupta2020precision}.
Moreover, low-cost accelerometers typically operate within a range of ±2g or ±8g, which easily leads to sensor saturation, data loss, and signal distortion in high-dynamic environments. For example, in industrial automation, accurately capturing complex motions requires accelerometers with a range of up to ±16g \cite{niu2022accelerometer}. However, industrial-grade accelerometers meeting these requirements are priced between \$10 and \$20 per unit, which is too expensive for consumer-end deployment. Top-tier accelerometers, like Xsens, can exceed \$1500 per unit due to their superior accuracy and range. In comparison, these widely used low-cost accelerometers are available for as little as \$0.20 to \$0.50 per unit. While affordable, these sensors are inadequate for many demanding applications. In healthcare, low-range accelerometers struggle to detect rapid movements or sudden events, such as falls, where accelerations often surpass 10g, causing missed alerts and defective health monitoring. Therefore, extending the range of low-cost accelerometers is crucial for enabling broader and more robust applications in various realms. In summary, by using advanced algorithms to enhance low-cost sensors, we can achieve high-end performance at a fraction of the cost, offering transformative potential across industries and making cutting-edge experiences accessible to all.


The rapid development of artificial intelligence \cite{xu2024shortform,wang2020adaptive,liu2024endogaussian}, particularly in generative deep learning models \cite{xu2024gaussianstego,wang2024chinese}, offers promising avenues for enhancing the range and accuracy of low-cost accelerometers. A potential solution lies in training a generative model to map low-cost sensor signals to high-cost ones, thereby improving the quality of low-cost signals.
However, it is impractical to obtain frame-by-frame paired data between low-cost and high-cost sensors \cite{pei2023markov}. Unpaired data pose significant challenges in training generative models due to the lack of supervision and guidance \cite{li2024endora}, often resulting in unreliable generated signals that fail to meet the stringent requirements of accelerometer applications. To address the severe problem in training with unpaired data, we propose a HEROS-GAN, which leverages optimal transport theory to maximize the use of supervisory information from unpaired data and injects Laplacian energy into the generator to encourage local changes. The core contributions of this paper are encapsulated in the following four aspects.

\begin{itemize}
    \item We propose to utilize deep learning algorithms for extending accelerometers range for the first time and introduce generative deep learning methods into accelerometer signal processing.
    \item Considering the lack of supervision from unpaired data, we design an Optimal Transport Supervision (OTS) to explore potential correlations within unpaired data, providing the model with as much supervisory information as possible.
    \item We design a Modulated Laplace Energy for GANs, guiding the model to generate more reasonable local changes, thereby enriching the generated signal details and breaking range limitation.
    \item We release the first accelerometer signal enhancement dataset in Github. Based on this dataset, our HEROS-GAN can extend the range of low-cost accelerometer signals from 8g to 16g, while reducing signal noise by an order of magnitude.
\end{itemize}

\section{Related Work}
\subsection{Over-Range Signal Recovery for Accelerometers}
Restoring over-range signals for accelerometers is an unexplored area within the current research landscape. Over-range signals are lost when the acceleration of the measured object exceeds the sensor's range, presenting a critical challenge in high dynamic applications such as automotive crash testing, industrial machinery monitoring, and sports science. However, few attempts are dedicated to the accelerometer over-range signal restoration problem.
Several factors contribute to the scarcity of research in this area. The nonlinearity and complexity of signal distortion when an accelerometer exceeds its measurement range make the restoration task highly challenging. Also, the recovery of over-range signals is highly context-dependent and relies on experience observing massive data, which is difficult for traditional non-data-driven models \cite{yang2024visionzip}. Finally, obtaining paired high-range and low-range data for training data-driven models is inherently tricky. The rarity of such paired datasets impedes the development of supervised learning approaches, further complicating the task of over-range signal restoration. Consequently, most existing efforts have focused on improving the dynamic range of sensors through hardware advancements rather than algorithm design \cite{wu2024design}.

\subsection{Signal Quality Enhancement for Accelerometers}
Compared to the rarely explored area of over-range signal restoration, considerable research has focused on reducing noise in accelerometer signals.
Traditional signal denoising approaches rely on various filtering techniques, including Kalman filters, Savitzky-Golay filters \cite{8713728}, and empirical mode decomposition \cite{liu2020denoising}. While these methods have proven effective in separating the noise from the signal and improving signal quality, they usually rely on prior knowledge of the signal or noise characteristics \cite{skog2009car}, causing poor generalization ability.
In contrast, data-driven methods learn the denoising function from data without relying on the information about signal characteristics, which can adapt to different sensors and noise patterns.
In data-driven approaches, generative deep learning models directly map low-cost signals to high-cost signals, providing a more effective way to enhance signal quality. However, their superior performance typically relies on strictly paired training data, which is impractical to obtain for low-cost and high-cost accelerometers \cite{wu2019survey}.
Therefore, few studies have attempted to apply GANs to improve accelerometer signals. 

\begin{figure*}[t]
	\centering
	\includegraphics[width=0.95\textwidth,height=0.32\textheight]{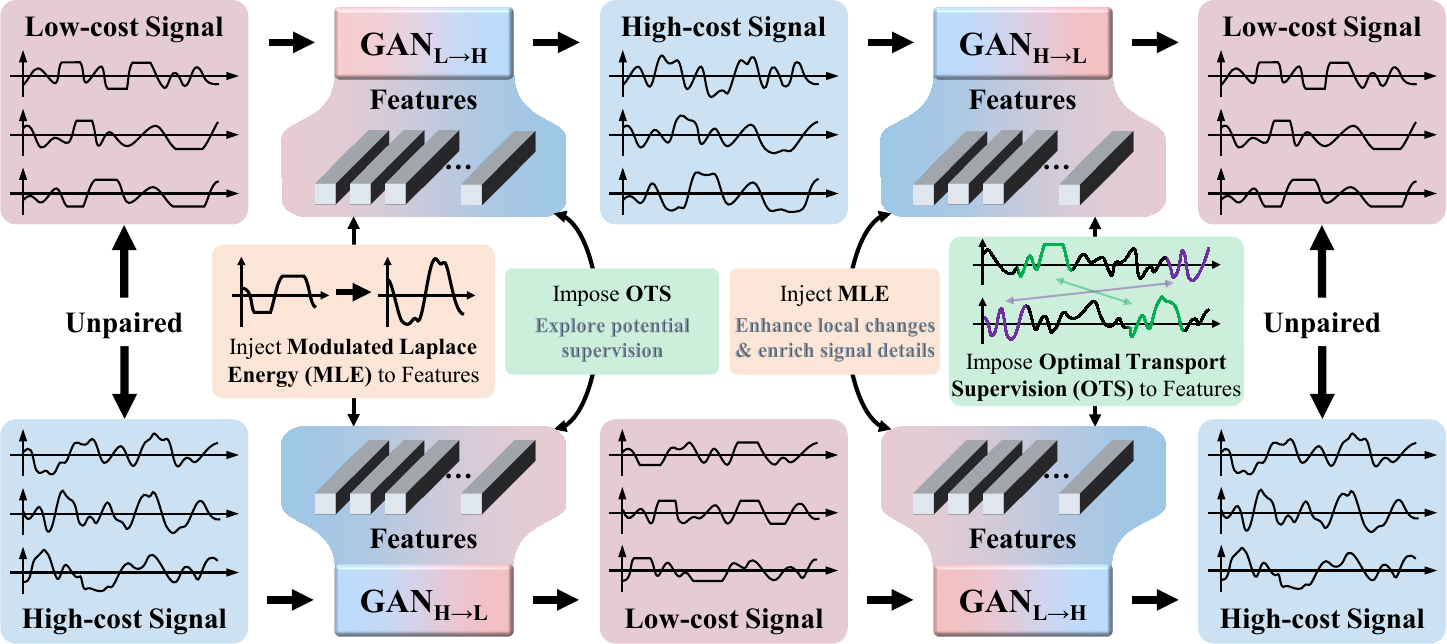} 
	\caption{Architecture of the HEROS-GAN. MLE (orange) and OTS (green) are applied to feature interaction on both sides.}
	\label{Framework of HEROS-GAN}
\end{figure*}
\section{Methodology}
The primary challenge in enhancing low-cost accelerometer signals lies in the inability to obtain strictly paired high-cost and low-cost sensor signals. Consequently, end-to-end methods with fully supervised training are impractical for converting low-quality signals into high-quality ones. To address this, we utilize CycleGAN as the baseline to construct a mapping between unpaired signals of varying qualities. Given the lack of paired data for guidance and supervision, we propose a honed-energy regularized and optimal supervised GAN (HEROS-GAN), as illustrated in Fig. \ref{Framework of HEROS-GAN}. In this architecture, Optimal Transport Supervision (OTS) is designed to mine supervisory information from unpaired data, while Modulated Laplace Energy (MLE) guides the model to generate realistic local changes, thereby enriching signal details.

\subsection{Optimal Transport Supervision}
Although both high-cost and low-cost signals are fed into the network simultaneously, the absence of paired data precludes the use of simple element-wise constraints (such as L1 or L2 loss), to enforce low-cost signal features approaching high-cost ones. Nonetheless, unpaired data still exhibit similar characteristics within their feature layers \cite{yang2021univerapprocnn}. To exploit these similarities, we propose the Optimal Transport Supervision (OTS) mechanism to align the features of low-cost signals with those of high-cost signals, thereby maximizing the supervision information obtained from unpaired data.

\begin{figure}[h]
	\centering
	\includegraphics[width=1.0\linewidth]{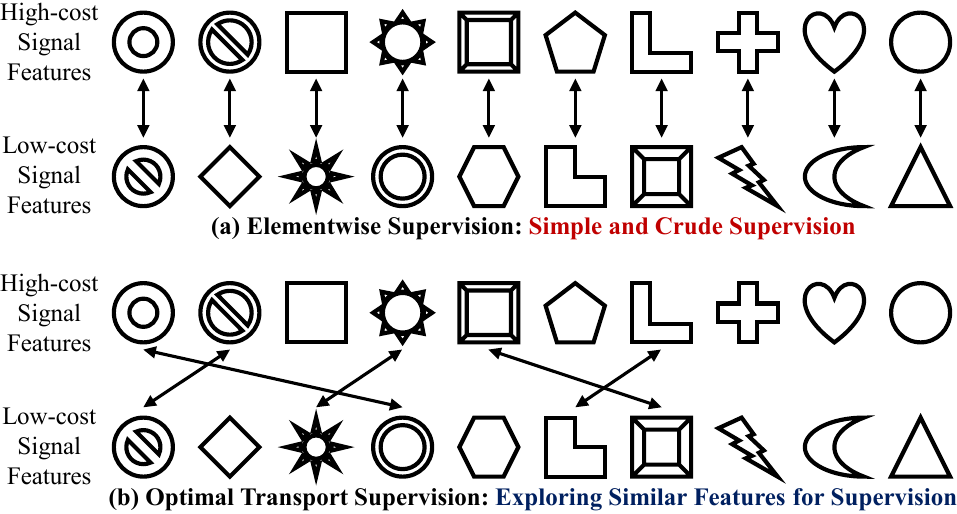} 
	\caption{Illustration of the Optimal Transport Supervision.}
	\label{OTS Mechanism}
\end{figure}

\begin{figure*}[t]
	\centering
	\includegraphics[width=1\textwidth]{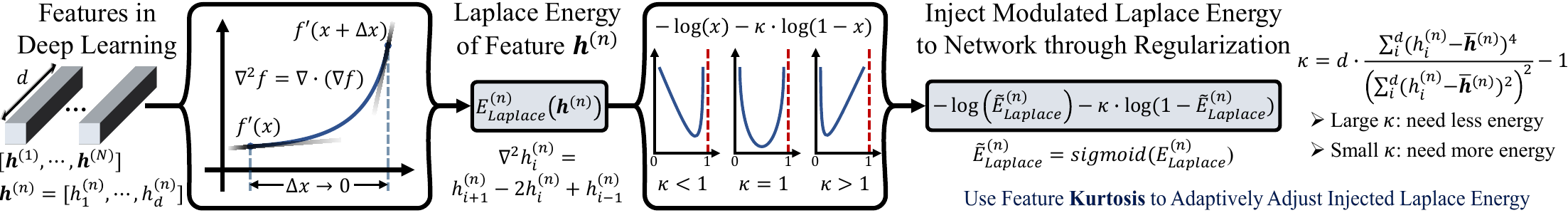} 
	\caption{Diagram of the Modulated Laplace Energy. The MLE computes Laplace energy from feature volatility and applies modulation through a regularization term, thereby controlling the volatility by adjusting the energy of the feature layers.}
	\label{MLE Framework}
\end{figure*}

As shown in Fig. \ref{OTS Mechanism}, the core idea of OTS is to explore similar features hidden in unpaired high-cost and low-cost signals for supervision, which is achieved by employing the optimal transport theory. Let $ F_L(x_L) = \{ f_{L1}, f_{L2}, \ldots, f_{LN} \} \in \mathbb{R}^{N \times d} $ represent the features of low-cost signals, and $ F_H(x_H) = \{ f_{H1}, f_{H2}, \ldots, f_{HN} \} \in \mathbb{R}^{N \times d} $ represent the features of high-cost signals. The goal of the optimal transport problem is to find an optimal mapping between these two feature distributions that minimizes the transportation cost. This problem can be formulated as:
\begin{equation}
	\min_{\gamma \in \Gamma(F_L, F_H)} \int_{\mathcal{X} \times \mathcal{Y}} c(f_{Li}, f_{Hj}) \, d\gamma(f_{Li}, f_{Hj})
	\label{optimal transport theory}
\end{equation}
where $\Gamma(F_L, F_H)$ represents the set of all joint distributions that couple $ F_L $ and $ F_H $, and $ c(f_{Li}, f_{Hj}) $ is the cost function, which is defined as $c(f_{Li}, f_{Hj}) = e^{1 - {f_{Li} \cdot f_{Hj}}}$.
The Sinkhorn algorithm can be applied to Eq. \ref{optimal transport theory} for approximating the optimal transport solution, which provides a transport mapping \( T: \mathcal{X} \to \mathcal{Y} \) by mining the feature consistency between low-cost and high-cost signals. This mapping allows us to identify and align the most similar features between the two distributions. Subsequently, we can define the OTS loss $\mathcal{L}_{OTS}$ to impose appropriate supervision that encourages low-cost signal features to align with certain high-cost signal features.
\begin{equation}
	\label{Loss_OTS}
	\begin{aligned}
		\mathcal{L}_{OTS} & = \mathbb{E}_{x_L \sim P_L, x_H \sim P_H} \big[ \|F_{G_L}(x_L) - T(F_H(x_H))\|^2 \\
		& + \|F_{G_H}(x_H) - T^{-1}(F_L(x_L))\|^2 \big]
	\end{aligned}
\end{equation}
where $P_L$ and $P_H$ denote the data distributions for domains of low-cost and high-cost signals. By minimizing $\mathcal{L}_{OTS}$, the generative model is encouraged to produce low-cost signal features that are aligned with the high-cost signal features. 




\subsection{Modulated Laplace Energy}

Enhancing the detail of generated signals is crucial for accurately capturing the subtle variations inherent in accelerometer data. Low-cost signals tend to be less sensitive, resulting in less detailed and more simplified signals \cite{yuan2024self}. In contrast, high-cost accelerometers can capture richer details and finer variations due to their higher sensitivity. This disparity directly affects performance and reliability in applications such as health monitoring, mobile device interactions, and precision engineering. Traditional GANs often struggle to generate these fine-grained details, leading to oversimplified and less realistic signals \cite{ijcai2024p567}.
To address this issue, we leverage the Laplacian operator, denoted as $\nabla^2$, which measures the second-order derivatives of a function and is effective in highlighting fine details and variations within the signal. Based on Laplacian operator, we propose the concept of Laplace Energy for features in a deep learning architecture, as shown in Equation \ref{equ E_laplace} and \ref{equ Laplace}:
\begin{equation}
	\label{equ E_laplace}
		E_{Laplace}^{(n)}(h^{(n)}) = \sum_{i=2}^{d-1} (\nabla^2 h_i^{(n)})^2
\end{equation}
\begin{equation}
	\label{equ Laplace}
	\nabla^2 h_i^{(n)} = h_{i+1}^{(n)}-2h_{i}^{(n)}+h_{i-1}^{(n)}
\end{equation}
where $ h^{(n)} \in \mathbb{R}^{d} $ denotes the $n$-th feature within the network, $d$ is the dimension of the feature, and $h_{i}^{(n)}$ is the $i$-th element in the feature $h^{(n)}$. The Laplace Energy evaluates the volatility of the features, with higher Laplace Energy indicating stronger volatility and more significant local variations, while lower energy suggests smoother and less detailed features in the generated signal. An intuitive approach is to design a regularization term based on Laplace Energy to guide the model in enhancing the feature volatility. However, unrestricted enhancement of Laplace Energy is clearly unreasonable, as it could introduce noise and instability in the generated signals. Therefore, we devise a Modulated Laplace Energy (MLE) as shown in Fig. \ref{MLE Framework}, which adaptively injects an appropriate amount of energy into the feature via regulation term $R_{MLE}$.
\begin{equation}
	\label{Loss_OTS}
	R_{MLE} = -\log \left({E}_{Laplace}\right) - \kappa \cdot \log \left(1 - {E}_{Laplace}\right)
\end{equation}
\begin{equation}
	\label{Loss_OTS}
	\tilde{E}_{Laplace} = \sigma({E}_{Laplace})
\end{equation}
where $\sigma$ denotes the sigmoid function, normalizing the Laplace energy to the range $(0, 1)$. $\kappa$ is a modulation parameter, set to feature kurtosis that is shifted to range $(0, +\infty)$.

This regularization term exhibits important properties. Firstly, when $\tilde{E}_{Laplace}$ approaches 0 or 1, \( R_{MLE} \) tends towards infinity, imposing a strong penalty on both very low and very high Laplace energy values. 
So the Laplace energy should remain within a moderate range. For instance, when \( \kappa = 1 \), the regularization term is symmetric, and its minimum occurs at $\tilde{E}_{Laplace} = 0.5$, thereby avoiding the injection of excessive or insufficient energy.

Secondly, the modulation parameter $\kappa$ controls the amount of injected energy by controlling the minimum point of the regularization term.
Specifically, high kurtosis (large \( \kappa \)) indicates strong volatility in the feature, necessitating less energy. Under this scenario, a large \( \kappa \) exactly ensures that the minimum of \( R_{MLE} \) occurs at a lower value of $\tilde{E}_{Laplace}$, leading to lower Laplace energy. 
Conversely, low kurtosis (small \( \kappa \)) indicates weak volatility, necessitating more energy. Under this scenario, a small \( \kappa \) exactly shifts the minimum of \( R_{MLE} \) towards a higher value of $\tilde{E}_{Laplace}$, resulting in higher Laplace energy.
This adaptive behavior ensures that the MLE mechanism injects the appropriate amount of energy into features based on their volatility characteristics. High-kurtosis features, which are naturally more fluctuating, receive less added energy, while low-kurtosis features, which are naturally smoother, receive more energy to enhance their details. 
By minimizing \( R_{MLE} \), the model generates finely detailed and stable signals, providing effective guidance for generative models in the absence of supervised information.

\section{Experiments and Results}
\subsection{Experiment Dataset}
Given the absence of a dedicated dataset for accelerometer over-range signals, we created a Low-cost Accelerometer Signal Enhancement Dataset (LASED) using 10 different smartphones equipped with built-in accelerometers, which are representative low-cost accelerometers \cite{jimenez2009comparison}. The specifications of the smartphones and their internal sensors are detailed in Table \ref{IMU specification}, where the cost of these sensors does not exceed \$0.5.
To evaluate our model's robustness, we utilize only one type of smartphone for collecting training data, while the remaining nine smartphones are exclusively used for testing. This setup imposes a significant challenge, requiring the model to generalize across varying hardware specifications, thereby rigorously assessing its performance.
Moreover, an eight-camera optical equipment (Nokov Mars2H) is employed to assist in motion capture.	
During the data collection process, each smartphone is subjected to vigorous shaking in multiple directions to induce signal overload that exceeded the measurement range of the accelerometers. This simulation replicates real-world scenarios where accelerometers encounter high dynamic forces and output overload signals. 
All experiments are implemented by Pytorch with NVIDIA RTX 4090 GPU and Intel(R) Xeon Gold 6330 CPU.
\begin{table}[h]
	\centering
	\setlength{\tabcolsep}{1.2mm}
	{
		\resizebox{\linewidth}{!}{
			\begin{tabular}{c|c|c|c|c}
				\toprule
				\multicolumn{1}{c|}{Dataset} & Smartphone & Release &\multicolumn{1}{c|}{Sensor} & \multicolumn{1}{c}{Unit price} \\
				\midrule
				\multicolumn{1}{c|}{Training} & HONOR Magic 4 & Feb 2022 &LSM6DSR	&\$0.35 \\
				\midrule
				\multicolumn{1}{c|}{\multirow{9}[13]{*}{Testing}} & HUAWEI P40 & Mar 2020 &LSM6DSM	&\$0.30 \\
				\cmidrule{2-5}          & OPPO Reno 6 &May 2021 & ICM-40607	&\$0.28 \\
				\cmidrule{2-5}          & HUAWEI P40 Pro & Apr 2020 &LSM6DSO	&\$0.33 \\
				\cmidrule{2-5}          & Realme GT &Mar 2021 & BMI160	&\$0.21 \\
				\cmidrule{2-5}          & Xiaomi 11 &Dec 2020 & BHI260AB	&\$0.30 \\
				\cmidrule{2-5}          & Lenovo Legion Phone & Aug 2020 &ICM-42605	&\$0.20  \\
				\cmidrule{2-5}          & VIVO T2x & May 2022 &LSM6DSO	&\$0.33	\\
				\cmidrule{2-5}          & iPhone 13 & Sep 2021 &Undisclosed	&/  \\
				\cmidrule{2-5}          & iPhone 12 & Oct 2020 &Undisclosed	&/  \\
				\bottomrule
			\end{tabular}%
		}
	}
	\caption{The built-in sensor of some smartphones. 
		}
	\label{IMU specification}%
\end{table}%

\begin{table*}[!h]
	\centering
	\setlength{\tabcolsep}{1.2mm}
	{
		\resizebox{\linewidth}{!}{
			\begin{tabular}{c|c|c|c|c|c|c|c|c}
				\toprule
				\multicolumn{3}{c|}{\multirow{2}[3]{*}{Architecture}} & \multicolumn{2}{c|}{CSRE$_{\tau}$ / g ($9.8m/s^2$)} & \multirow{2}[2]{*}{$ZVRE$ / $m/s$} & \multicolumn{3}{c}{Allan Variance Analysis}\\
				\cmidrule{4-5} \cmidrule{7-9}   \multicolumn{3}{c|}{} & ${\tau=15g}$ & ${\tau=6g}$ & 
				\multicolumn{1}{c|}{}  & \multicolumn{1}{c|}{QN} & \multicolumn{1}{c|}{VRW} & \multicolumn{1}{c}{BI} \\
				\midrule
				\multicolumn{3}{c|}{Raw signal (No processing)} & \multicolumn{1}{c|}{0.562} & \multicolumn{1}{c|}{1.981} & \multicolumn{1}{c|}{350.78} & \multicolumn{1}{c|}{1.205} & \multicolumn{1}{c|}{1.917} & \multicolumn{1}{c}{3.119} \\
				\midrule
				\multicolumn{1}{c|}{\multirow{3}[4]{*}{\makecell{Model\\Driven}}} & \multirow{3}[4]{*}{\makecell{Non\\Machine\\Learning}} & EMD-Kalman filter  & 0.388 (-31.0\%) & 1.871 (-5.6\%) & 355.65 (+1.4\%) & 0.491 (-59.3\%) & 0.576 (-70.0\%) & 0.643 (-79.4\%) \\
				\cmidrule{3-9}          & \multicolumn{1}{c|}{} & Savitzky Golay filter & 0.436 (-22.4\%) & 1.845 (-6.9\%) & 340.22 (-3.0\%) & 0.542 (-55.0\%) & 0.706 (-63.2\%) & 0.76 (-75.6\%) \\
				\cmidrule{3-9}          & \multicolumn{1}{c|}{} & Matlab 2023 & \underline{0.321 (-42.9\%)} & 1.779 (-10.2\%) & 361.12 (+3.0\%) & 1.211 (+0.5\%) & 1.906 (-0.6\%) & 3.109 (-0.3\%) \\
				\midrule
				\multicolumn{1}{c|}{\multirow{6}[10]{*}{\makecell{Data\\Driven}}} & \multirow{4}[7]{*}{\makecell{Fully\\Supervised}} & CNN-o & 0.922 (+64.1\%) & \underline{1.519 (-23.3\%)} & 294.96 (-15.9\%) & 1.306 (+8.4\%) & 2.235 (+16.6\%) & 3.437 (+10.2\%) \\
				\cmidrule{3-9}          & \multicolumn{1}{c|}{} & GRU-LSTM & 0.791 (+40.8\%) & 1.742 (-12.1\%) & 317.78 (-9.4\%) & 1.297 (+7.6\%) & 2.139 (+11.6\%) & 3.531 (+13.2\%) \\
				\cmidrule{3-9}          & \multicolumn{1}{c|}{} & GRU-LSTM-o & 0.848 (50.9\%) & 1.587 (-19.9\%) & \underline{310.4 (-11.5\%)} & 1.375 (+14.1\%) & 2.203 (+14.9\%) & 3.746 (+20.1\%) \\
				\cmidrule{3-9}          & \multicolumn{1}{c|}{} & kNN  & 0.739 (+31.5\%) & 1.697 (-14.3\%) & 339.83 (-3.1\%) & 0.988 (-18.0\%) & 1.545 (-19.4\%) & 2.717 (-12.9\%) \\
				\cmidrule{2-9}          & Unsupervised & IMUDB & 0.523 (-6.9\%) & 1.632 (-17.6\%) & 378.33 (+7.9\%) & \underline{0.189 (-84.3\%)} & \underline{0.274 (-85.7\%)} & \underline{0.444 (-85.8\%)} \\
				\cmidrule{2-9}          & \makecell{Weakly\\Supervised} & HEROS-GAN (Ours) & \textbf{0.199 (-64.6\%)} & \textbf{0.329 (-83.4\%)} & \textbf{21.05 (-94.0\%)} & \textbf{0.057 (-95.3\%)} & \textbf{0.06 (-96.9\%)} & \textbf{0.065 (-97.9\%)} \\
				\bottomrule
			\end{tabular}%
		}
	}
	\caption{Comparison of latest methods. Considering the dimension difference of indicators, we give the error reduction ratio relative to the raw signal in parentheses for convenient comparison. We bold the best and underline the 2nd best results.}
	\label{tab:comparison}%
\end{table*}%

\subsection{Evaluation Metrics}
As the study of generating over-range accelerometer signals is relatively unexplored, there is a lack of metrics to evaluate the accuracy of the generated signals. We, therefore, propose two metrics, Clipped Signal Reconstruction Error (CSRE) and Zero-Velocity Residual Error (ZVRE), to assess the effectiveness of over-range reconstruction methods.
For CSRE, the high-cost sensor signal \( S_{hc}(t) \) is artificially clipped at multiple thresholds \( \tau \) to simulate the signal saturation phenomenon of low-cost sensors with different ranges, resulting in clipped signals \( S_{clip, \tau}(t) \).  
The reconstruction method is then applied to these clipped signals, producing reconstructed signals \( S_{recon, \tau}(t) \). CSRE is calculated by comparing the reconstructed signals with the original unclipped signals by the formula:
\begin{equation}
	\label{Loss_OTS}
	\text{CSRE}_{\tau} = \sqrt{\frac{1}{N} \sum_{t=1}^{N} (S_{hc}(t) - S_{recon, \tau}(t))^2}
\end{equation}
where \( N \) is the total number of time frames.

ZVRE measures the physical plausibility of generated accelerometer signals. The physical property that the integral of the acceleration signal over transition period from rest to vigorous shaking and back to rest should be zero for each axis (x, y, and z) is essential for validating the physical accuracy of the generated signals. The ZVRE is calculated as the absolute value of the integrated acceleration, indicating the deviation from the expected zero velocity.
\begin{equation}
	\text{ZVRE}_\text{axis} = \left| \int_{0}^{T} a_{axis}(t) \, dt \right|
\end{equation}
where \( T \) denotes the time period. This metric is just zero when the reconstructed signals maintain the zero-velocity condition after periods of motion, thereby validating their physical plausibility. 
In addition to the proposed CSRE and ZVRE, we employ Allan variance \cite{pei2023markov} to evaluate the accuracy of acceleration signals under static conditions. Allan variance is a classical time-domain technique that provides the quantitative indicators of sensor signal quality, including quantization noise (QN), velocity random walk (VRW), and bias instability (BI).

\subsection{Comparative Results}

Recent acceleration denoising methods, including optimized CNN (CNN-o) \cite{chen2022towards}, GRU-LSTM \cite{han2021hybrid}, optimized GRU-LSTM (GRU-LSTM-o) \cite{boronakhin2022optimization}, and kNN \cite{engelsman2023data}, rely on fully supervised training by frame-paired high-cost and low-cost signals. However, collecting such paired data under high-dynamic conditions is impractical. To the end, we used clipped and unclipped high-cost signals to train these methods for generating over-range and high-quality signals. 
Additionally, MATLAB versions released after 2023 introduced an over-range signal reconstruction function based on polynomial fitting techniques, which is one of the few existing methods that directly tackle the issue of signal saturation. Consequently, we included it in our comparative study. Moreover, we introduced some of the latest signal enhancement methods applicable to IMUs for comparison \cite{yuan2023simple}. All comparative methods were implemented strictly following the procedures described in their papers or using their open-source codes. The results are shown in Table \ref{tab:comparison}.

\begin{figure}[t]
	\centering
	\includegraphics[width=1.0\linewidth,height=0.33\textheight]{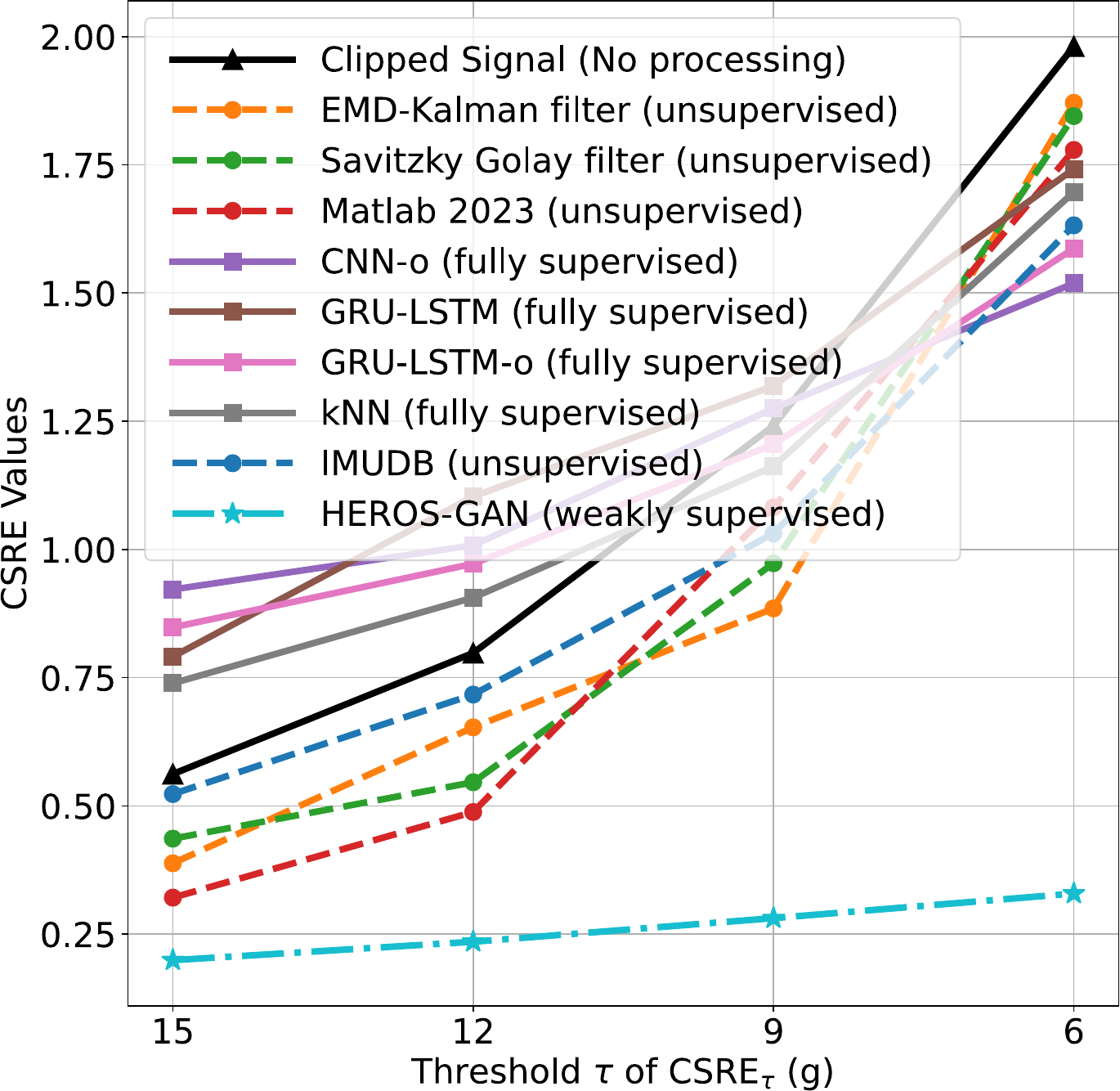} 
	\caption{Visualization of CSRE for different methods as the clipping threshold $\tau$ decreases from $15g$ to $6g$. 
	}
	\label{Visualization CSRE}
\end{figure}

The Clipped Signal Reconstruction Error is tested on clipped high-cost signals. When the clipping is minimal, such as at \(\tau = 15g\), model-driven methods perform well since they only need to fit the small clipped portions without altering the original signal. In contrast, deep learning models (except ours) tend to modify the entire input signal, occasionally resulting in a higher CSRE. However, as the clipping level increases, traditional model-driven methods struggle. They fail to effectively reconstruct larger clipped sections due to their simplistic fitting approach, leading to significant reconstruction errors. To illustrate this, we visualized the CSRE for different methods as \(\tau\) decreased from 15g to 6g, as shown in Fig. \ref{Visualization CSRE}.
It is evident that the CSRE of the proposed HEROS-GAN is much lower than all comparative methods, attributed to the identity loss. This loss preserves the non-clipped parts of the input high-cost signal, enabling accurate reconstruction of the clipped sections without altering the original signal structure, thus handling both minimal and severe clipping scenarios.

\begin{figure}[t]
	\centering
	\includegraphics[width=1.0\linewidth,height=0.27\textheight]{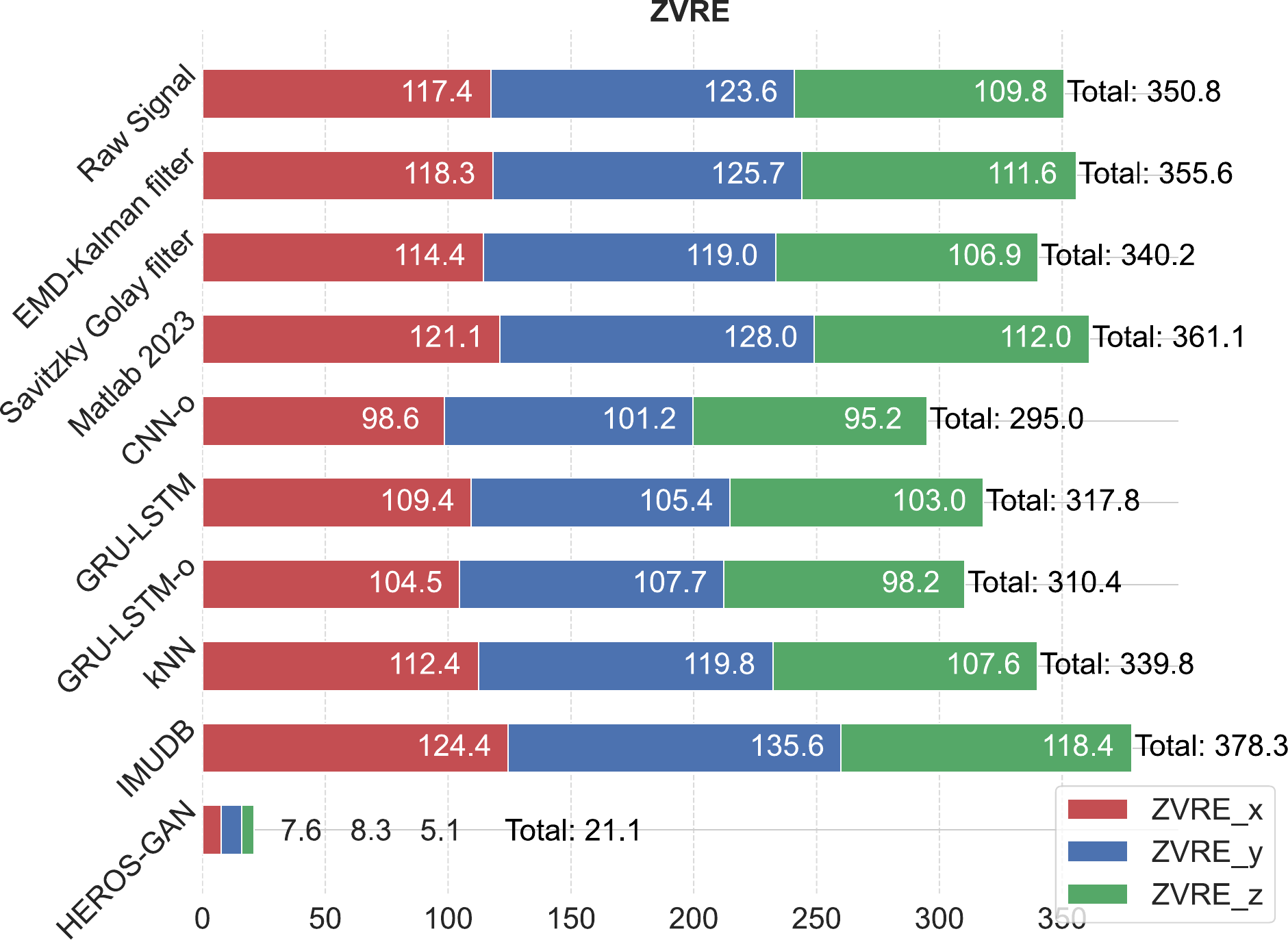} 
	\caption{Visualization of the ZVRE for different methods and the raw low-cost signal across the x, y, and z axes. 
	}
	\label{Visualization ZVRE}
\end{figure}

The Zero-Velocity Residual Error is tested on saturated low-cost signals. Fully supervised methods, trained with high-cost signals, attempt to emulate high-quality signal characteristics, achieving slight reductions in the ZVRE of the test low-cost signals. However, their lack of exposure to low-cost signals limits their effectiveness.
In contrast, unsupervised learning methods trained with low-cost signals and model-driven methods never observe high-cost sensor signals, leaving them unaware of the physical plausibility, which causes significant velocity deviations after integration, making these methods ineffective for ZVRE reduction.
To demonstrate this, we visualize the ZVRE of the x, y, and z axes for all comparative methods and the raw low-cost signal, as shown in Fig. \ref{Visualization ZVRE}. It can be observed that the HEROS-GAN framework outperforms all comparative methods due to its ability to observe both low-cost and high-cost signals. Despite unpaired signals, our Optimal Transport Supervision mechanism exploits their potential correlations to provide the maximum supervisory information. Consequently, the enhanced signal accurately complies with physical laws, ensuring minimal velocity deviation after the Static-Dynamic-Static process.

Allan variance analysis evaluates the accuracy of static low-cost signals. Fully supervised methods trained on paired clipped and non-clipped high-cost signals do not perform well in this context. Since these methods have never encountered low-cost signals during training, they cannot effectively denoise and may even degrade the signal quality.
Traditional model-driven methods exhibit some denoising capability. However, their limited fitting ability restricts their effectiveness.
The unsupervised IMUDB method achieves notable denoising results, significantly reducing Allan variance metrics. However, its lack of exposure to high-cost signals prevents full exploitation of superior signal characteristics, resulting in suboptimal denoising performance.
Our HEROS-GAN framework exhibits exceptional denoising capability, achieving the lowest Allan variance metrics and significantly improving static signal quality, thereby setting a new standard for accelerometer signal enhancement.

\subsection{Ablation Study}
CycleGAN, which utilizes both low-cost and high-cost signals, performs relatively well in static evaluations since the static signal segments could be paired. However, its performance in CSRE and ZVRE is not well, indicating that modeling the mapping between unpaired data is challenging for CycleGAN. The MLE and OTS, respectively, enhance signal quality, each excelling in their own aspects. MLE is particularly effective in generating over-range signals, resulting in a lower CSRE$_{\tau=6g}$ (0.474) due to the ability of MLE to adaptively adjust the feature energy, i.e., increasing energy under high-dynamic conditions to overcome range limitations while decreasing energy under stable conditions to reduce noise. OTS excels in simulating high-cost signal characteristics to elevate inferior signals, leading to lower ZVRE, QN, VRW, and BI. Replacing OTS with L1 supervision, which forces unpaired signals into strict alignment, results in significantly poor performance, underscoring  the necessity of flexible and adaptive supervision mechanisms like OTS. The HEROS-GAN combining MLE and OTS outperforms SOTA methods by an order of magnitude across all metrics.

\begin{table}[htbp]
	\centering
	\setlength{\tabcolsep}{1.9mm}
	{
		\resizebox{\linewidth}{!}{
	\begin{tabular}{c|c|c|c|c|c}
		\toprule
		Architecture & \multicolumn{1}{c|}{CSRE$_{\tau=6g}$} & \multicolumn{1}{c|}{ZVRE} & \multicolumn{1}{c|}{QN} & \multicolumn{1}{c|}{VRW} & \multicolumn{1}{c}{BI} \\
				\midrule
		w/o all (CycleGAN) & 1.803 & 332.8 & 0.166 & 0.180  & 0.193 \\
		\midrule
		w/ MLE & \underline{0.474} & 195.7 & 0.103 & 0.136 & 0.159  \\
		\midrule
		w/ OT & 0.588 & \underline{39.41} & \underline{0.059} & \underline{0.072}  & \underline{0.079} \\
		\midrule
		w/ L1 & 2.091 & 433.9 & 1.337 & 2.532 & 3.796 \\
		\midrule
		\makecell{w/ MLE+OTS\\(HEROS-GAN)} & \textbf{0.329} & \textbf{21.05} & \textbf{0.057} & \textbf{0.060}  & \textbf{0.065} \\
		\bottomrule
	\end{tabular}%
}}
	\caption{Ablation experiments on the proposed modules.}
	\label{tab:ablation}%
\end{table}%

\section{Discussion}
The proposed OTS module leverages optimal transport theory to explore potential consistencies between unpaired or weakly-paired data, providing as much supervisory information as possible for generative models and breaking the limitations of strictly paired data. The core idea of OTS lies in aligning features across different data distributions. In multimodal learning, OTS can align feature distributions across modalities, facilitating cross-modal generation, translation, and enhancement tasks. In domain adaptation, OTS can align features from different domains. In medical imaging analysis, where data distributions from different imaging devices (e.g., MRI and CT) vary significantly \cite{li2022domain}, OTS can align the feature distributions across devices, enabling diagnostic model transfer between devices. In robotic perception, OTS can align features from heterogeneous sensors or platforms, supporting multi-platform collaboration and knowledge transfer. 
In tasks such as remote sensing image processing, unsupervised representation learning, and cross-modal retrieval, the feature alignment capability of OTS can also play a critical role.
Moreover, its ability to model optimal mappings between distributions could be extended to decision-making systems, such as reinforcement learning, where it could optimize the transfer of policies between agents operating in different environments.
In summary, OTS provides soft yet strict supervision for weakly paired or even unpaired data and features, unlocking hidden consistencies between distributions and uncovering meaningful relationships.

The design of the MLE regularization term offers a powerful and flexible approach for controlling model behavior, with a wide range of potential applications in deep learning tasks. Its formulation can be simplified as \(-\log(x) - \alpha \cdot \log(1-x)\), which serves as a loss function to penalize extreme outputs in deep learning models. More importantly, the parameter \(\alpha \in (0, +\infty)\) endows the model with the ability to regulate its output preferences. When \(\alpha\) approaches 0, as the regularization term decreases, the model tends to output values close to 1; conversely, when \(\alpha\) approaches \(+\infty\), the model’s output tends towards 0 as the regularization term decreases. By setting \(\alpha\) to reflect specific model characteristics, we can softly guide the model’s behavior, avoiding overly rigid constraints.
For instance, in this work, \(\alpha\) is set to the kurtosis of the signal features, enabling the model to inject an appropriate amount of Laplacian energy, neither too much nor too little. 
In tasks like feature alignment or semantic consistency constraints, a suitable \(\alpha\) can be employed to softly regulate the similarity between features, facilitating more natural alignment across different modalities or tasks. Additionally, this regularization term can also be applied to soft adversarial training, where adjusting \(\alpha\) helps balance the model, preventing it from becoming overly biased towards a specific direction.
In threshold selection or hyperparameter tuning, it allows the model to adaptively adjust thresholds or hyperparameters based on its intrinsic characteristics by the preference factor $\alpha$. 
In summary, this regularization term, with its capability for soft constraints, preserves flexibility while guiding the model's learning process, demonstrating significant versatility and utility.

\section{Conclusion}

The widespread usage of low-cost accelerometers is often limited by their accuracy and range. Due to the lack of paired data, few studies utilize generative models to transfer low-cost accelerometer signals into high-cost equivalents. This paper proposes a HEROS-GAN for signal accuracy and range enhancement based on unpaired data of independently measured low-cost and high-cost accelerometer signals, which integrates with the Optimal Transport Supervision (OTS) and the Modulated Laplace Energy (MLE) modules. The OTS module leverages the optimal transport theory to explore optimal mapping between such unpaired data, thereby maximizing supervisory information. 
We provide a rigorous mathematical proof to ensure the existence of this optimal mapping and demonstrate that it can converge during training. Moreover, we mathematically prove that OTS can reduce the instability and oscillations of GAN, overcoming one of the key problems for GAN architecture.
The MLE module calculates and adaptively adjusts the Laplace energy of features within the generator, promoting local changes and enriching signal details. Considering the absence of dedicated datasets, we release a Low-cost Accelerometer Signal Enhancement Dataset (LASED) in GitHub, providing the first data experimental platform for accelerometer range and accuracy enhancement. We also devise two metrics, Clipped Signal Reconstruction Error (CSRE) and Zero-Velocity Residual Error (ZVRE), to assess the accuracy and physical plausibility of the enhanced signals.
Experimental results demonstrate that a CycleGAN combined with either OTS or MLE alone outperforms current SOTA methods in accelerometer signal enhancement with a tenfold improvement. The HEROS-GAN framework, integrating both OTS and MLE, achieves exceptional results, effectively doubling the accelerometer range while reducing signal noise by two orders of magnitude.
The physical plausibility of the generated signals (evidenced by the low ZVRE) confirms the practical applicability of HEROS-GAN and sets a standard in acceleration signal processing.

\section*{Acknowledgments}
This work was supported by the Science Center Program of National Natural Science Foundation of China under Grant 62188101, the Natural Science Foundation of Guangdong under Grant 2020A1515010812 and 2021A1515011594, and China Scholarship Council under Grant 202306120304. 
We sincerely appreciate the Education Center of Experiments and Innovations (Analysis and Testing Center) at Harbin Institute of Technology, Shenzhen, for their support. Furthermore, we sincerely appreciate the help provided by Professor Ji Hui from the National University of Singapore.

\bibliographystyle{named}
\bibliography{reference}

\begin{thebibliography}{37}
\providecommand{\natexlab}[1]{#1}

\bibitem[{Boronakhin et~al.(2022)Boronakhin, Shalymov, Larionov, Khanh, and
  Yen}]{boronakhin2022optimization}
Boronakhin, A.~M.; Shalymov, R.~V.; Larionov, D.~Y.; Khanh, N.~Q.; and Yen,
  N.~T. 2022.
\newblock Optimization of an Inertial Sensor De-Noising Method using a Hybrid
  Deep Learning Algorithm.
\newblock In \emph{2022 Conference of Russian Young Researchers in Electrical
  and Electronic Engineering (ElConRus)}, 1335--1340. IEEE.

\bibitem[{Caesar et~al.(2020)Caesar, Bankiti, Lang, Vora, Liong, Xu, Krishnan,
  Pan, Baldan, and Beijbom}]{caesar2020nuscenes}
Caesar, H.; Bankiti, V.; Lang, A.~H.; Vora, S.; Liong, V.~E.; Xu, Q.; Krishnan,
  A.; Pan, Y.; Baldan, G.; and Beijbom, O. 2020.
\newblock nuscenes: A multimodal dataset for autonomous driving.
\newblock In \emph{Proceedings of the IEEE/CVF conference on computer vision
  and pattern recognition}, 11621--11631.

\bibitem[{Chen, Taha, and Chodavarapu(2022)}]{chen2022towards}
Chen, H.; Taha, T.~M.; and Chodavarapu, V.~P. 2022.
\newblock Towards improved inertial navigation by reducing errors using deep
  learning methodology.
\newblock \emph{Applied Sciences}, 12(7): 3645.

\bibitem[{Ehatisham-ul Haq et~al.(2021)Ehatisham-ul Haq, Arsalan, Raheel, and
  Anwar}]{ehatisham2021expert}
Ehatisham-ul Haq, M.; Arsalan, A.; Raheel, A.; and Anwar, S.~M. 2021.
\newblock Expert-novice classification of mobile game player using smartphone
  inertial sensors.
\newblock \emph{Expert Systems with Applications}, 174: 114700.

\bibitem[{Engelsman and Klein(2023)}]{engelsman2023data}
Engelsman, D.; and Klein, I. 2023.
\newblock Data-driven denoising of stationary accelerometer signals.
\newblock \emph{Measurement}, 218: 113218.

\bibitem[{Gromov et~al.(2019)Gromov, Abbate, Gambardella, and Giusti}]{8794399}
Gromov, B.; Abbate, G.; Gambardella, L.~M.; and Giusti, A. 2019.
\newblock Proximity Human-Robot Interaction Using Pointing Gestures and a
  Wrist-mounted IMU.
\newblock In \emph{2019 International Conference on Robotics and Automation
  (ICRA)}, 8084--8091.

\bibitem[{Gupta et~al.(2020)Gupta, Moghimi, Jeong, Gupta, Inan, and
  Ayazi}]{gupta2020precision}
Gupta, P.; Moghimi, M.~J.; Jeong, Y.; Gupta, D.; Inan, O.~T.; and Ayazi, F.
  2020.
\newblock Precision wearable accelerometer contact microphones for longitudinal
  monitoring of mechano-acoustic cardiopulmonary signals.
\newblock \emph{NPJ digital medicine}, 3(1): 19.

\bibitem[{Han et~al.(2021)Han, Meng, Zhang, and Yan}]{han2021hybrid}
Han, S.; Meng, Z.; Zhang, X.; and Yan, Y. 2021.
\newblock Hybrid deep recurrent neural networks for noise reduction of MEMS-IMU
  with static and dynamic conditions.
\newblock \emph{Micromachines}, 12(2): 214.

\bibitem[{Jimenez et~al.(2009)Jimenez, Seco, Prieto, and
  Guevara}]{jimenez2009comparison}
Jimenez, A.~R.; Seco, F.; Prieto, C.; and Guevara, J. 2009.
\newblock A comparison of pedestrian dead-reckoning algorithms using a low-cost
  MEMS IMU.
\newblock In \emph{2009 IEEE International Symposium on Intelligent Signal
  Processing}, 37--42. IEEE.

\bibitem[{Karaim, Noureldin, and Karamat(2019)}]{8713728}
Karaim, M.; Noureldin, A.; and Karamat, T.~B. 2019.
\newblock Low-cost IMU Data Denoising using Savitzky-Golay Filters.
\newblock In \emph{2019 International Conference on Communications, Signal
  Processing, and their Applications (ICCSPA)}, 1--5.

\bibitem[{Li et~al.(2022{\natexlab{a}})Li, Lin, Mao, Lin, Qi, Ding, Huang,
  Liang, and Yu}]{li2022domain}
Li, C.; Lin, X.; Mao, Y.; Lin, W.; Qi, Q.; Ding, X.; Huang, Y.; Liang, D.; and
  Yu, Y. 2022{\natexlab{a}}.
\newblock Domain generalization on medical imaging classification using
  episodic training with task augmentation.
\newblock \emph{Computers in biology and medicine}, 141: 105144.

\bibitem[{Li et~al.(2024{\natexlab{a}})Li, Liu, Fan, Li, Liu, Pan, and
  Yuan}]{xu2024gaussianstego}
Li, C.; Liu, H.; Fan, Z.; Li, W.; Liu, Y.; Pan, P.; and Yuan, Y.
  2024{\natexlab{a}}.
\newblock Gaussianstego: A generalizable stenography pipeline for generative 3d
  gaussians splatting.
\newblock \emph{arXiv preprint arXiv:2407.01301}.

\bibitem[{Li et~al.(2024{\natexlab{b}})Li, Liu, Liu, Feng, Li, Liu, Chen, Shao,
  and Yuan}]{li2024endora}
Li, C.; Liu, H.; Liu, Y.; Feng, B.~Y.; Li, W.; Liu, X.; Chen, Z.; Shao, J.; and
  Yuan, Y. 2024{\natexlab{b}}.
\newblock Endora: Video generation models as endoscopy simulators.
\newblock In \emph{International Conference on Medical Image Computing and
  Computer-Assisted Intervention}, 230--240. Springer Nature Switzerland Cham.

\bibitem[{Li et~al.(2023)Li, Zhang, Jin, Hu, and Wang}]{li2023attitude}
Li, P.; Zhang, W.-A.; Jin, Y.; Hu, Z.; and Wang, L. 2023.
\newblock Attitude Estimation using Iterative Indirect Kalman with Neural
  Network for Inertial Sensors.
\newblock \emph{IEEE Transactions on Instrumentation and Measurement}.

\bibitem[{Li et~al.(2022{\natexlab{b}})Li, Chen, Niu, Zhuang, Gao, Hu, and
  El-Sheimy}]{9500152}
Li, Y.; Chen, R.; Niu, X.; Zhuang, Y.; Gao, Z.; Hu, X.; and El-Sheimy, N.
  2022{\natexlab{b}}.
\newblock Inertial Sensing Meets Machine Learning: Opportunity or Challenge?
\newblock \emph{IEEE Transactions on Intelligent Transportation Systems},
  23(8): 9995--10011.

\bibitem[{Liu et~al.(2020{\natexlab{a}})Liu, Zhang, Zhang, and
  Zhu}]{liu2020wearable}
Liu, S.; Zhang, J.; Zhang, Y.; and Zhu, R. 2020{\natexlab{a}}.
\newblock A wearable motion capture device able to detect dynamic motion of
  human limbs.
\newblock \emph{Nature communications}, 11(1): 5615.

\bibitem[{Liu et~al.(2020{\natexlab{b}})Liu, Caruso, Ilg, Dong, Mourikis,
  Daniilidis, Kumar, and Engel}]{liu2020tlio}
Liu, W.; Caruso, D.; Ilg, E.; Dong, J.; Mourikis, A.~I.; Daniilidis, K.; Kumar,
  V.; and Engel, J. 2020{\natexlab{b}}.
\newblock Tlio: Tight learned inertial odometry.
\newblock \emph{IEEE Robotics and Automation Letters}, 5(4): 5653--5660.

\bibitem[{Liu et~al.(2020{\natexlab{c}})Liu, Chen, Wei, Yang, and
  Xing}]{liu2020denoising}
Liu, Y.; Chen, G.; Wei, Z.; Yang, J.; and Xing, D. 2020{\natexlab{c}}.
\newblock Denoising method of MEMS gyroscope based on interval empirical mode
  decomposition.
\newblock \emph{Mathematical Problems in Engineering}, 2020: 1--12.

\bibitem[{Liu et~al.(2024)Liu, Li, Yang, and Yuan}]{liu2024endogaussian}
Liu, Y.; Li, C.; Yang, C.; and Yuan, Y. 2024.
\newblock EndoGaussian: Real-time Gaussian Splatting for Dynamic Endoscopic
  Scene Reconstruction.
\newblock \emph{arXiv preprint arXiv:2401.12561}.

\bibitem[{Malayappan et~al.(2022)Malayappan, Lakshmi, Rao, Ramaswamy, and
  Pattnaik}]{malayappan2022sensing}
Malayappan, B.; Lakshmi, U.~P.; Rao, B.~P.; Ramaswamy, K.; and Pattnaik, P.~K.
  2022.
\newblock Sensing techniques and interrogation methods in optical MEMS
  accelerometers: A review.
\newblock \emph{IEEE Sensors Journal}, 22(7): 6232--6246.

\bibitem[{Niu et~al.(2022)Niu, Sheng, Gao, and Zhou}]{niu2022accelerometer}
Niu, Y.; Sheng, L.; Gao, M.; and Zhou, D. 2022.
\newblock Accelerometer fault detection for rotary steerable drilling tool
  systems under strong noises.
\newblock \emph{IEEE Transactions on Instrumentation and Measurement}, 71:
  1--11.

\bibitem[{Pei et~al.(2023)Pei, Fan, Du, Zhang, Yuan, and Quan}]{pei2023markov}
Pei, H.; Fan, W.; Du, P.; Zhang, K.; Yuan, L.; and Quan, W. 2023.
\newblock Markov noise in atomic spin gyroscopes: Analysis and suppression
  based on allan deviation.
\newblock \emph{IEEE Transactions on Instrumentation and Measurement}, 72:
  1--9.

\bibitem[{Skog and Handel(2009)}]{skog2009car}
Skog, I.; and Handel, P. 2009.
\newblock In-car positioning and navigation technologies—A survey.
\newblock \emph{IEEE Transactions on Intelligent Transportation Systems},
  10(1): 4--21.

\bibitem[{Wang et~al.(2020)Wang, Wang, Hu, Liu, and Zhao}]{wang2020adaptive}
Wang, Y.; Wang, Y.; Hu, G.; Liu, Y.; and Zhao, Y. 2020.
\newblock Adaptive skewness kurtosis neural network: enabling communication
  between neural nodes within a layer.
\newblock In \emph{International Conference on Neural Information Processing},
  498--507. Springer.

\bibitem[{Wang, Xu, and Zhao(2024)}]{10614868}
Wang, Y.; Xu, J.; and Zhao, Y. 2024.
\newblock Wavelet Encoding Network for Inertial Signal Enhancement via Feature
  Supervision.
\newblock \emph{IEEE Transactions on Industrial Informatics}, 20(11):
  12924--12934.

\bibitem[{Wang and Zhao(2024{\natexlab{a}})}]{wang2024chinese}
Wang, Y.; and Zhao, Y. 2024{\natexlab{a}}.
\newblock Chinese Inertial GAN for Writing Signal Generation and Recognition.

\bibitem[{Wang and Zhao(2024{\natexlab{b}})}]{10318077}
Wang, Y.; and Zhao, Y. 2024{\natexlab{b}}.
\newblock Handwriting Recognition Under Natural Writing Habits Based on a
  Low-Cost Inertial Sensor.
\newblock \emph{IEEE Sensors Journal}, 24(1): 995--1005.

\bibitem[{Wang and Zhao(2024{\natexlab{c}})}]{ijcai2024p567}
Wang, Y.; and Zhao, Y. 2024{\natexlab{c}}.
\newblock Scale and Direction Guided GAN for Inertial Sensor Signal
  Enhancement.
\newblock In \emph{Proceedings of the Thirty-Third International Joint
  Conference on Artificial Intelligence, {IJCAI-24}}, 5126--5134.
\newblock Main Track.

\bibitem[{Wang and Zhao(2024{\natexlab{d}})}]{wang2024wavelet}
Wang, Y.; and Zhao, Y. 2024{\natexlab{d}}.
\newblock Wavelet Dynamic Selection Network for Inertial Sensor Signal
  Enhancement.
\newblock In \emph{Proceedings of the AAAI Conference on Artificial
  Intelligence}, volume~38, 15680--15688.

\bibitem[{Wu et~al.(2024)Wu, Yu, Zhang, Zhang, and Zhou}]{wu2024design}
Wu, Q.; Yu, Z.; Zhang, X.; Zhang, J.; and Zhou, T. 2024.
\newblock Design and Implementation of an Integrated Array Accelerometer with
  Expanded Dynamic Range Based on Adaptive Data Selection Fusion.
\newblock \emph{IEEE Transactions on Instrumentation and Measurement}.

\bibitem[{Wu et~al.(2019)Wu, Zhu, Du, and Tang}]{wu2019survey}
Wu, Y.; Zhu, H.-B.; Du, Q.-X.; and Tang, S.-M. 2019.
\newblock A survey of the research status of pedestrian dead reckoning systems
  based on inertial sensors.
\newblock \emph{International Journal of Automation and Computing}, 16: 65--83.

\bibitem[{Xu et~al.(2024)Xu, Zhou, Yang, and Li}]{xu2024shortform}
Xu, H.; Zhou, J.; Yang, M.; and Li, J. 2024.
\newblock Shortform ugc video quality assessment based on multi-level video
  fusion with rank-aware.
\newblock In \emph{Proceedings of the IEEE/CVF Conference on Computer Vision
  and Pattern Recognition Workshops}, volume~7.

\bibitem[{Yang et~al.(2024)Yang, Chen, Tian, Wang, Li, Yu, and
  Jia}]{yang2024visionzip}
Yang, S.; Chen, Y.; Tian, Z.; Wang, C.; Li, J.; Yu, B.; and Jia, J. 2024.
\newblock VisionZip: Longer is Better but Not Necessary in Vision Language
  Models.
\newblock \emph{arXiv preprint arXiv:2412.04467}.

\bibitem[{Yang et~al.(2023)Yang, Liu, Zhang, Pan, Guo, Li, Chen, Gao, Guo, and
  Zhang}]{yang2023lidar}
Yang, S.; Liu, J.; Zhang, R.; Pan, M.; Guo, Z.; Li, X.; Chen, Z.; Gao, P.; Guo,
  Y.; and Zhang, S. 2023.
\newblock Lidar-llm: Exploring the potential of large language models for 3d
  lidar understanding.
\newblock \emph{arXiv preprint arXiv:2312.14074}.

\bibitem[{Yang, Wang, and Yang(2021)}]{yang2021univerapprocnn}
Yang, Y.; Wang, Y.; and Yang, S. 2021.
\newblock A UniverApproCNN with Universal Approximation and Explicit Training
  Strategy.
\newblock In \emph{Collaborative Computing: Networking, Applications and
  Worksharing: 17th EAI International Conference, CollaborateCom 2021, Virtual
  Event, October 16-18, 2021, Proceedings, Part II 17}, 297--315. Springer.

\bibitem[{Yuan et~al.(2024)Yuan, Plekhanova, Walmsley, Reynolds, Maddison,
  Bucan, Gehrman, Rowlands, Ray, Bennett et~al.}]{yuan2024self}
Yuan, H.; Plekhanova, T.; Walmsley, R.; Reynolds, A.~C.; Maddison, K.~J.;
  Bucan, M.; Gehrman, P.; Rowlands, A.; Ray, D.~W.; Bennett, D.; et~al. 2024.
\newblock Self-supervised learning of accelerometer data provides new insights
  for sleep and its association with mortality.
\newblock \emph{NPJ digital medicine}, 7(1): 86.

\bibitem[{Yuan and Wang(2023)}]{yuan2023simple}
Yuan, K.; and Wang, Z.~J. 2023.
\newblock A simple self-supervised imu denoising method for inertial aided
  navigation.
\newblock \emph{IEEE Robotics and Automation Letters}, 8(2): 944--950.

\end{thebibliography}

\end{document}